\documentclass[a4paper, 10pt, conference]{ieeeconf}      

\IEEEoverridecommandlockouts                              

\usepackage{amsmath, amssymb}
\usepackage{graphicx}
\usepackage{booktabs}
\usepackage{multirow}
\usepackage{xcolor}
\usepackage{url}
\usepackage{caption}
\usepackage{subcaption}
\usepackage{algorithm}
\usepackage{algpseudocode}
\usepackage{tabularx}
\usepackage{array}
\usepackage{float}
\usepackage{threeparttable}
\title{\LARGE \bf
Drift-Aware Temporal Graph Rewiring (DATGR) for Adaptive Semantic Modeling in Biomedical Text
}

\author{%
Bharathwaj Vijayakumar$^{1}$ and Sahana K. Varadaraju$^{2}$%
\thanks{$^{1}$Bharathwaj Vijayakumar is with Institutional Data and Analytics, Rowan University, NJ, USA. 
        (e-mail: vijayakumar@rowan.edu)}%
\thanks{$^{2}$Sahana K. Varadaraju is with Information Resources \& Technology, Rowan University, NJ, USA.
        (e-mail: varadaraju@rowan.edu)}%
\thanks{© 2026 IEEE. Personal use of this material is permitted. Permission 
        from IEEE must be obtained for all other uses, in any current or future media, 
        including reprinting/republishing this material for advertising or promotional 
        purposes, creating new collective works, for resale or redistribution to 
        servers or lists, or reuse of any copyrighted component of this work in other 
        works. DOI: 10.1109/CAI68641.2026.11536228}%
}

\begin{document}

\maketitle
\thispagestyle{empty}
\pagestyle{empty}

\begin{abstract}
    Biomedical language evolves rapidly as new discoveries emerge, causing traditional text models to lose semantic fidelity over time. 
    Static embeddings and co-occurrence graphs cannot capture such evolution, leading to performance degradation in retrieval and knowledge discovery tasks. This paper introduces a \textbf{Drift-Aware Temporal Graph Rewiring (DATGR)} 
    framework that models concept evolution by dynamically updating co-occurrence edges based on estimated semantic drift. Instead of retraining embeddings for each time slice, DATGR performs lightweight, feedback-driven rewiring using a logistic update rule applied to edge weights. 
    Evaluated on the Biomedical Multi-Relation Corpus (BIOMRC), the method achieved a mean Area Under the Receiver Operating Characteristic (AUROC) improvement of approximately 0.066 absolute difference (\(0.699\) vs.\ \(0.633\)) over a static baseline. 
    Area Under the Precision-Recall Curve (AUPRC) remained comparable (\(0.738\) vs.\ \(0.744\)), showing that drift-aware adaptation enhances link-prediction recall without a loss in precision. 
    These results demonstrate that edge-level adaptation effectively captures temporal semantic change in evolving biomedical text while remaining computationally efficient and interpretable.
\end{abstract}

\begin{keywords}
Semantic drift, temporal graph learning, link prediction, biomedical text mining, adaptive embeddings, concept evolution, retrieval-augmented generation (RAG)
\end{keywords}

\section{INTRODUCTION}
The exponential growth of biomedical publications has transformed scientific text into a continuously evolving knowledge ecosystem. Thousands of new abstracts appear weekly in PubMed and related repositories, introducing novel concepts, drug-compound interactions, and terminologies that shift the semantic structure of the domain. Capturing such concept evolution is critical for applications in literature mining, knowledge graph curation, and retrieval-augmented generation. However, most graph-based or embedding-based text models still assume static relationships between words and entities, which quickly become outdated as meaning drifts over time.

Traditional co-occurrence graphs~\cite{church1990word} and word embeddings such as word2vec~\cite{mikolov2013efficient} or GloVe~\cite{pennington2014glove} treat semantics as stationary. In reality, biomedical terms like ``immune checkpoint,'' ``mRNA vaccine,'' or ``viral vector'' evolve as research advances, resulting in measurable semantic drift~\cite{hamilton2016diachronic, tahmasebi2021survey, nicholson2023semantic}. Static models therefore misrepresent emerging associations and reduce retrieval accuracy when used for downstream analysis.

To mitigate this, several temporal embedding frameworks~\cite{bamler2017dynamic, nicholson2023semantic} and dynamic graph neural networks (DGNNs), and surveys thereof~\cite{kazemi2020dynamic,yuan2023continual}, have been proposed. While effective, many methods require retraining or sequential fine-tuning for each time slice; these approaches incur significant computational cost and limiting interpretability. Furthermore, most focus on node-level drift, how individual term vectors change, rather than on edge-level drift, which captures the evolving co-occurrence structure among terms.
Instead of retraining embeddings, DATGR rewires the word-co-occurrence graph using a drift-weighted logistic update rule, enabling continuous adaptation without full re-embedding. Evaluated on the \textit{BIOMRC} corpus, the proposed model achieved an absolute AUROC improvement of approximately 0.066 (\(0.699\) vs.\ \(0.633\)) over a static baseline, while maintaining comparable AUPRC performance (\(0.738\) vs.\ \(0.744\)). These results demonstrate that lightweight, feedback-driven edge updates can effectively capture semantic evolution while preserving precision.

The key contributions of this work are as follows:
\begin{itemize}
    \item A drift-aware rewiring mechanism for temporal co-occurrence graphs that models edge evolution efficiently without retraining embeddings (DATGR);
    \item An empirical demonstration on real biomedical text, showing improved link-prediction AUROC and stable precision-recall behavior across temporal windows;
    \item A scalable foundation for adaptive semantic modeling applicable to evolving domains such as biomedical knowledge tracking, retrieval augmentation, and ontology maintenance.
\end{itemize}

\section{RELATED WORK}
Research on modeling temporal change in language has evolved along three major directions: static co-occurrence graphs, temporal word embeddings, and dynamic graph learning.

Early approaches based on word co-occurrence statistics treated term relations as static~\cite{church1990word}.
While such representations capture word associations effectively at a single point in time, they ignore how meaning shifts as domains evolve. Distributional embeddings such as word2vec~\cite{mikolov2013efficient} and GloVe~\cite{pennington2014glove} improved semantic representation but remained time-invariant.
Recent work on temporal embeddings for biomedical literature
has explored how contextual meaning changes across time~\cite{zhou2022tde}.
To address temporal variability, researchers proposed diachronic or dynamic embeddings that explicitly model meaning change across time. Hamilton \textit{et al.}~\cite{hamilton2016diachronic} analyzed word embedding trajectories to uncover statistical laws of semantic change, while Tahmasebi \textit{et al.}~\cite{tahmasebi2021survey} surveyed computational methods for detecting lexical semantic change. Bamler and Mandt~\cite{bamler2017dynamic} introduced probabilistic dynamic embeddings, modeling word transitions as latent trajectories through time. 
Nicholson \textit{et al.}~\cite{nicholson2023semantic} analyzed large-scale biomedical corpora and demonstrated domain-specific semantic drift in biomedical scientific language.

Dynamic graph learning addresses temporal networks directly. Recent surveys of DGNNs and temporal knowledge graphs (e.g.,~\cite{kazemi2020dynamic}) summarize architectures that update node representations as edges arrive ~\cite{kazemi2020dynamic}. However, many such methods emphasize node embeddings rather than direct edge updates reflecting co-occurrence strength. The proposed DATGR differs by explicitly modeling edge drift via a lightweight logistic rewiring rule, achieving temporal sensitivity while remaining interpretable and efficient.

\section{METHODOLOGY}
\begin{figure}[!tbh]
    \centering
    \includegraphics[width=0.48\textwidth]{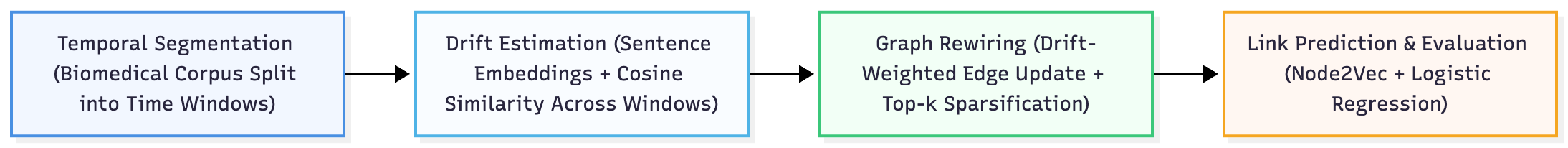}
    \caption{Overview of the DATGR (drift-aware temporal graph rewiring) framework, showing the stages of temporal segmentation, drift estimation, graph rewiring, and link-prediction evaluation.}
    \label{fig:system}
\end{figure}

The proposed DATGR framework models the evolution of biomedical semantics through incremental, edge-level graph adaptation rather than full re-embedding. It comprises four modules: (1) temporal segmentation of the corpus, (2) drift estimation using sentence embeddings, (3) drift-weighted graph rewiring, and (4) link prediction with node embeddings. Fig.~\ref{fig:system} illustrates this pipeline.

\subsection{Temporal Segmentation and Graph Construction}
The biomedical corpus is divided into $T$ chronological windows $\{W_1, W_2, \dots, W_T\}$, each representing a distinct semantic period. For every window, tokenized abstracts are converted into an undirected weighted graph $G_t=(V_t,E_t,W_t)$ where $V_t$ represents high-frequency biomedical terms and $E_t$ encodes co-occurrence within a context window of size $n$.
Each edge weight is normalized by the maximum observed frequency:
\begin{equation}
W_t(i,j) = \frac{\text{Count}(i,j)}{\max_{a,b}\text{Count}(a,b)}.
\label{eq:weight}
\end{equation}
This normalization allows comparability of edge strengths across windows with differing corpus sizes.

\subsection{Drift-Aware Rewiring Rule}
\noindent DATGR introduces a logistic edge update rule that adjusts edge weights between two terms $(i,j)$ using their prior weight, current co-occurrence strength, and local drift.

We denote by $s_t(i,j)=W_t(i,j)$ the normalized co-occurrence strength in the current window $t$ (cf. Eq.~\eqref{eq:weight}), and define the pairwise change
\begin{equation}
\Delta_{ij} \;=\; s_t(i,j) \;-\; s_{t-1}(i,j).
\end{equation}
Given the drift magnitudes $D_t(\cdot)$, the edge update from $t{-}1$ to $t$ is
\begin{equation}
\begin{aligned}
\hat{W}_{t}(i,j) \;=\; &(1-\eta)\,W_{t-1}(i,j) \\
&+\; \eta\,\sigma\!\Big(
b_0 \;+\; b_1\,s_t(i,j) \;+\; b_2\,\Delta_{ij} \\
&\qquad\qquad\qquad\;\;+\; b_3\,[D_t(i)+D_t(j)]
\Big),
\end{aligned}
\label{eq:update}
\end{equation}
where $\sigma(x)=1/(1+e^{-x})$ is the logistic transform and $\eta\in[0,1]$ is the adaptation rate. The first term preserves inertia through a convex combination of  $W_{t-1}$, while the second term injects innovation using current evidence ($s_t$), its change ($\Delta_{ij}$), and the local drift signal ($D_t$). Because $W_{t-1}\in[0,1]$ and $\sigma(\cdot)\in(0,1)$, the convex combination guarantees $\hat{W}_t(i,j)\in[0,1]$.

\paragraph*{Practical notes} We found $\eta\!\in\![0.1,0.2]$ to balance stability and responsiveness. Coefficients $(b_0,b_1,b_2,b_3)$ can be tuned on a validation slice; setting $b_2{=}0$ reduces the rule to a drift and strength-weighted update when change terms are noisy. A small threshold on $D_t(\cdot)$ may be applied to ignore negligible drift.

\subsection{Top-k Sparsification and Stability}
After rewiring, each node retains only its top-k neighbors ranked by $\hat{W}_t(i,j)$. This enforces sparsity and mitigates noise propagation due to spurious term co-occurrences. Empirically, k=5 provided the best trade-off between stability and adaptability, yielding consistent graph density across time.

\subsection{Node Embedding and Link Prediction}
Updated graphs $\hat{G}_t$ are embedded via Node2Vec to derive latent representations $Z_u$ for every node $u$. Each candidate edge $(u,v)$ is represented as the element-wise product:
\begin{equation}
z_{uv} = Z_u \odot Z_v.
\end{equation}
A logistic regression classifier is trained to predict whether $(u,v)$ will appear in $G_{t+1}$, effectively testing the model's ability to anticipate new semantic associations.

\subsection{Computational Complexity}
Let $|E_t|$ denote the number of edges per window. DATGR's update complexity is $O(|E_t|)$ since each edge weight is modified independently, avoiding the $O(|V|^2)$ cost of re-embedding or backpropagation through time. This results in substantially lower computational cost than dynamic GNN approaches while preserving a comparable temporal sensitivity.

\subsection{Algorithm Summary}
\begin{algorithm}[H]
\caption{Drift-Aware Temporal Graph Rewiring (DATGR).}
\begin{algorithmic}[1]
\State Split corpus into windows $\{W_1,\dots,W_T\}$
\For{$t = 2$ to $T$}
    \State Construct co-occurrence graph $G_t$
    \State Compute term drift $D_t(w)$
    \For{each edge $(i,j)\in G_{t-1}$}
        \State Update weight $\hat{W}_t(i,j)$ using Eq.~\eqref{eq:update}
    \EndFor
    \State Retain top-k edges per node
\EndFor
\State Train Node2Vec on $\hat{G}_t$; predict edges in $G_{t+1}$
\end{algorithmic}
\end{algorithm}
\section{EXPERIMENTS AND RESULTS}

\subsection{Dataset and Temporal Segmentation}
Experiments used the \textit{BIOMRC} dataset~\cite{pappas2020biomrc}, divided into $T=4$ windows of approximately 1000 abstracts each, forming pseudo-chronological snapshots.

\subsection{Graph Construction}
For each window, the 400 most frequent biomedical terms formed $V_t$, yielding $\sim$400 nodes and 4k-5k edges. Co-occurrence weights followed Eq.~\eqref{eq:weight}.

\subsection{Semantic Drift Estimation}
Embeddings were produced with all-MiniLM-L6-v2. Drift between windows was computed using cosine distance between consecutive term representations.

\subsection{Baselines and Configuration}
A static graph baseline (fixed edges) was compared to DATGR with identical settings: $\eta=0.2$, $(b_0,b_1,b_2,b_3)=(0,3,2,1)$, top-k=5; Node2Vec dim=64, walk length=8, 20 walks/node, context window=5, seed=42.

\subsection{Quantitative Results}
DATGR and the static baseline were compared using link-prediction performance across consecutive temporal windows. Performance was measured via AUROC and AUPRC, averaged across three consecutive window transitions (0→1, 1→2, 2→3). Table~\ref{tab:main} summarizes the results.

\begin{table}[!tbh]
    \caption{Static vs. Drift-Aware (DATGR) Link Prediction Performance.}
    \label{tab:main}
    \begin{center}
    \setlength{\tabcolsep}{1pt}
    \renewcommand{\arraystretch}{1.05}
    \begin{tabular}{|c|c|c|c|c|}
    \hline
    \textbf{Win.} & \textbf{Static AUROC} & \textbf{DATGR AUROC} & \textbf{Static AUPRC} & \textbf{DATGR AUPRC} \\
    \hline
    0 & 0.643 & \textbf{0.698} & \textbf{0.748} & 0.725 \\
    1 & 0.630 & \textbf{0.700} & 0.745 & \textbf{0.746} \\
    2 & 0.627 & \textbf{0.698} & 0.739 & \textbf{0.743} \\
    \hline
    \textbf{Mean} & 0.633 & \textbf{0.699} & \textbf{0.744} & 0.738 \\
    \hline
    \end{tabular}
    \end{center}
    \vspace{-2pt}
\end{table}

\begin{figure}[!tbh]
    \centering
    \includegraphics[width=0.48\textwidth]{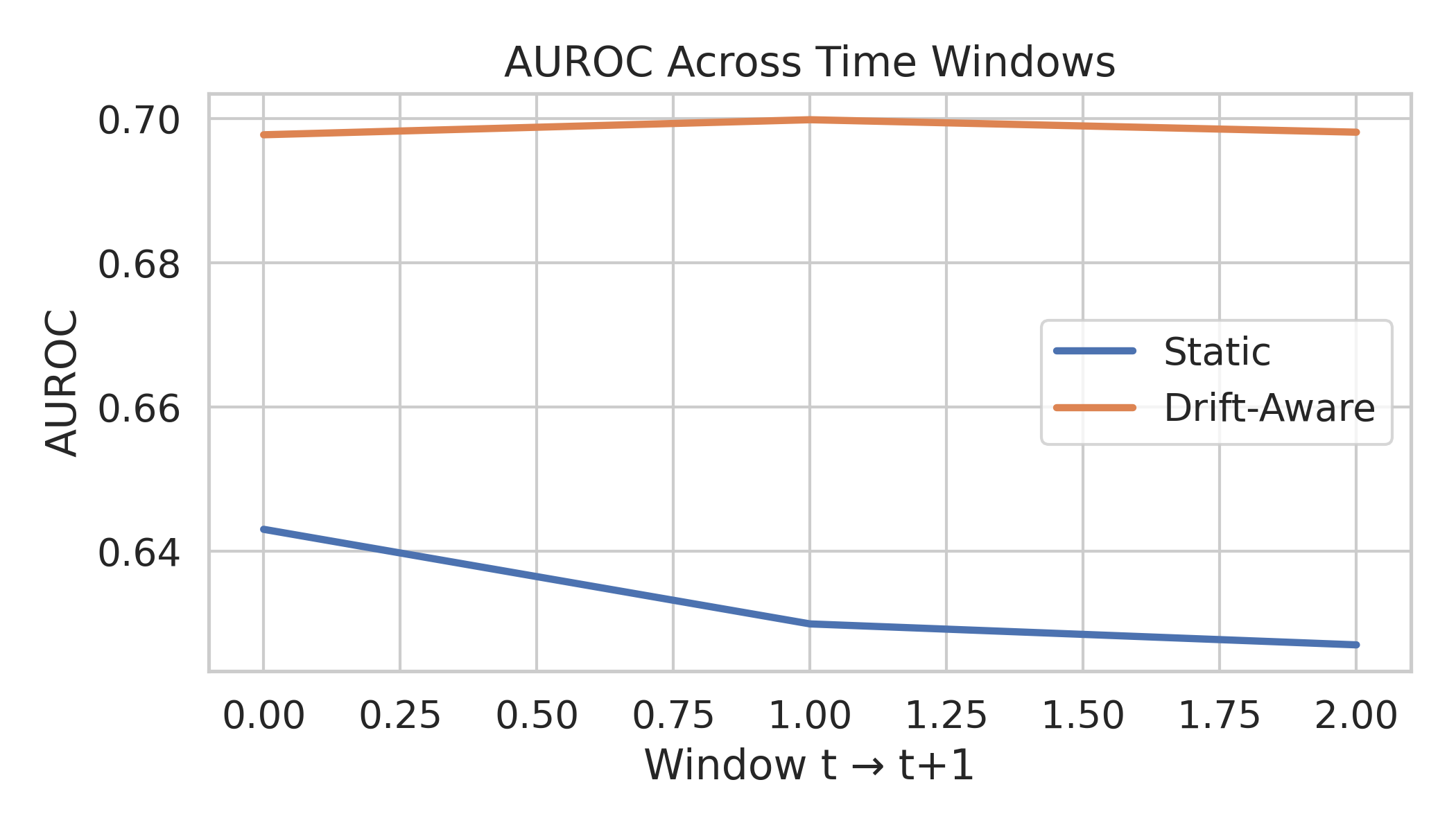}
    \caption{AUROC across temporal windows. DATGR consistently outperforms the static baseline, achieving gains of approximately 0.066 per window.}
    \label{fig:auroc}
\end{figure}

\begin{figure}[!tbh]
    \centering
    \includegraphics[width=0.48\textwidth]{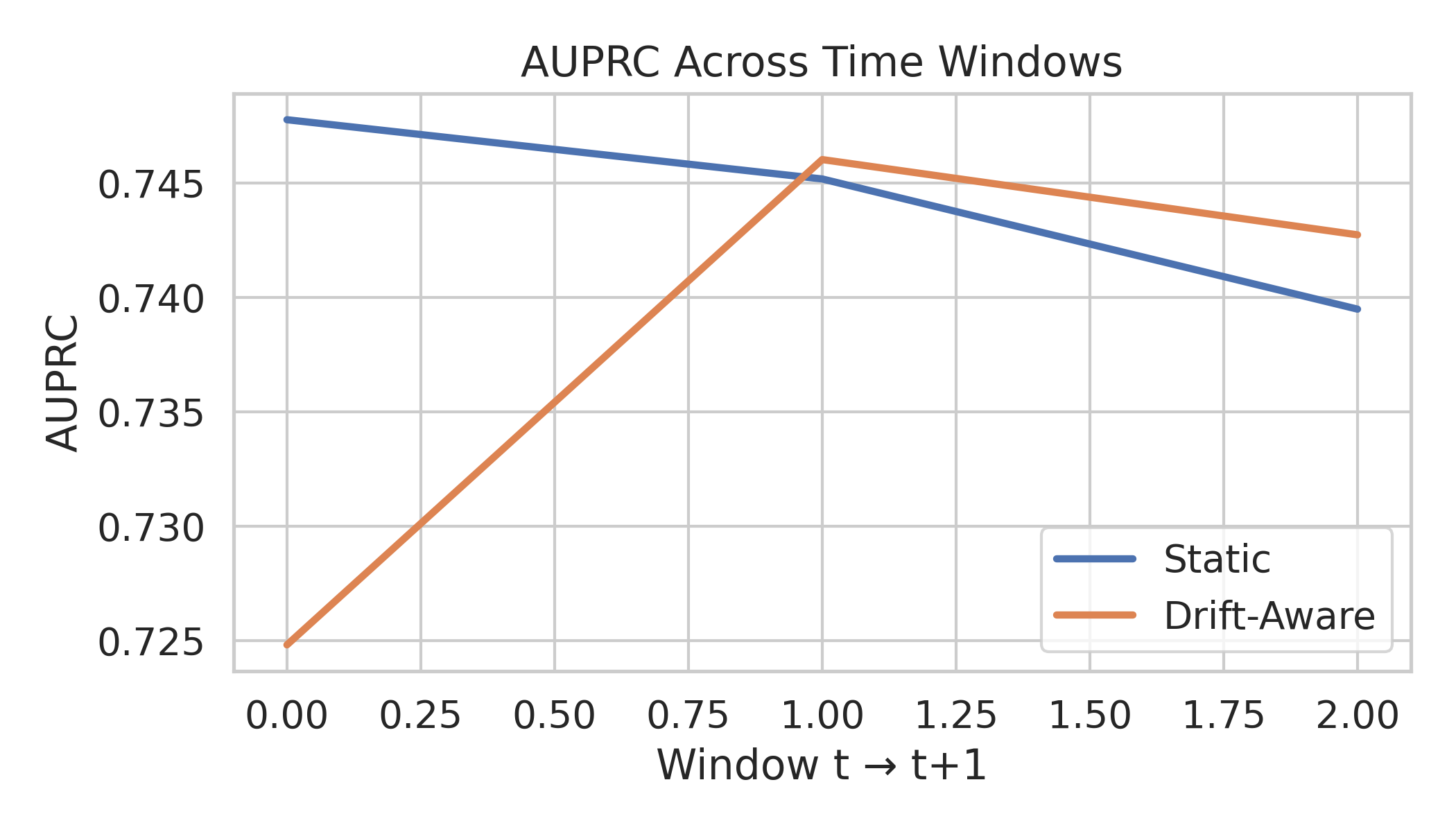}
    \caption{AUPRC across temporal windows. DATGR remains comparable to the static model, indicating improved recall without reducing precision.}
    \label{fig:auprc}
\end{figure}

\begin{figure}[!tbh]
    \centering
    \includegraphics[width=0.48\textwidth]{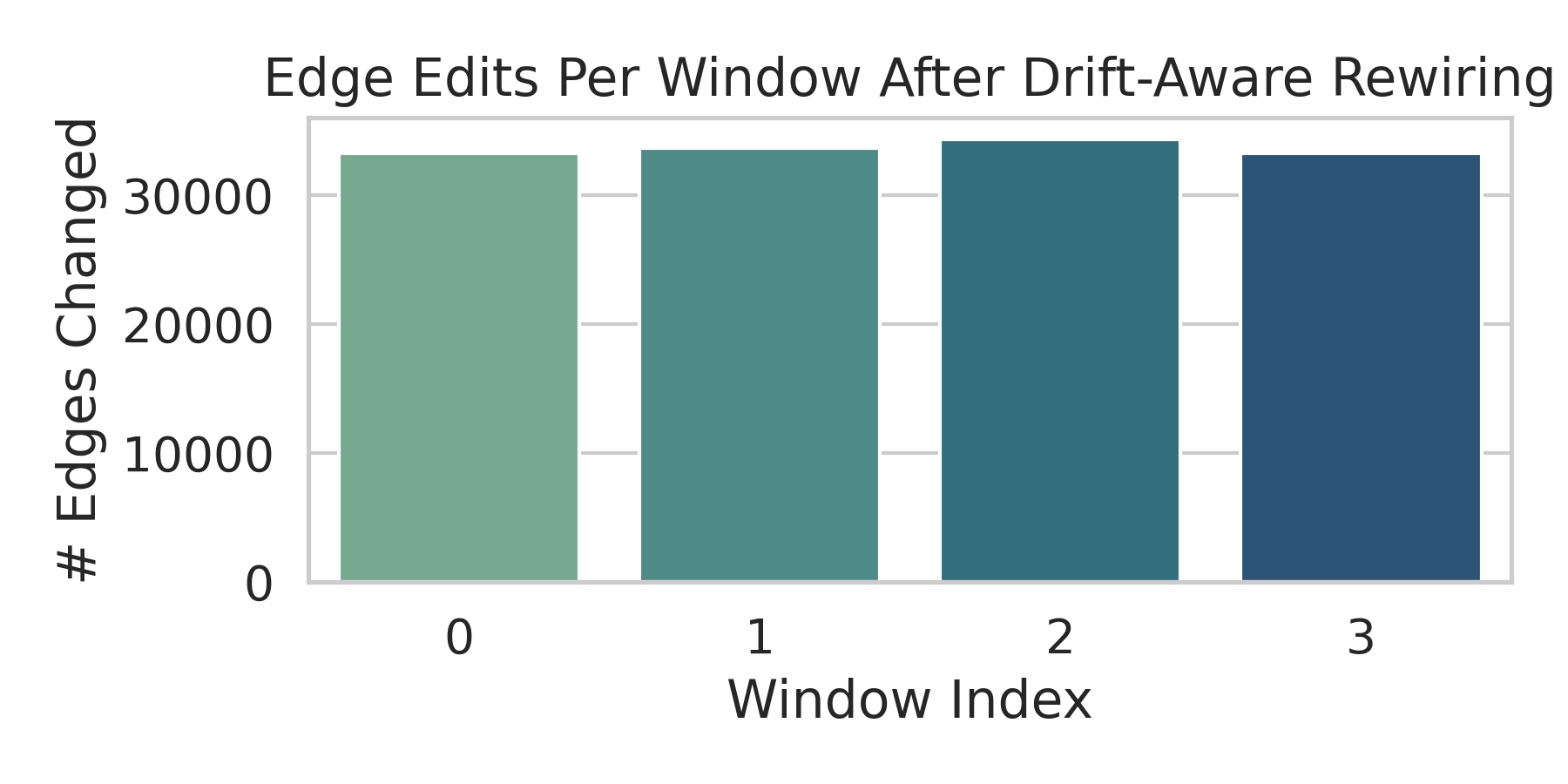}
    \caption{Edge edits per temporal window after DATGR rewiring, demonstrating stable yet adaptive structural updates.}
    \label{fig:edits}
\end{figure}

\subsection{Analysis of Figures and Results}
Fig.~\ref{fig:auroc} shows that DATGR achieves a consistent AUROC improvement across all windows, achieving an average absolute gain of approximately 0.066 compared to the static model. This indicates that our proposed method improves the model's ability to discriminate between emerging and obsolete semantic relations. The DATGR curve remains stable, reflecting robust generalization across time slices rather than overfitting to a single window.

In Fig.~\ref{fig:auprc}, both static and drift-aware models achieve similar AUPRC levels ($\sim0.74$). DATGR slightly surpasses the baseline in later windows, while remaining comparable overall. This stability demonstrates that DATGR preserves precision even as it adapts to new connections, suggesting balanced recall expansion.

Fig.~\ref{fig:edits} illustrates that DATGR modifies a consistent, moderate number of edges in each temporal transition, confirming that adaptation occurs without destabilizing the overall structure. This ``steady rewiring'' pattern aligns with the theoretical design: a small but meaningful number of edge updates per window efficiently encode semantic evolution while retaining the corpus's structural coherence.

Overall, these findings confirm that DATGR effectively balances precision and adaptability. Its gains in AUROC and stable AUPRC suggest a robust capacity to capture semantic drift while avoiding overfitting or excessive rewiring, key traits for scalable biomedical semantic modeling.

\section{Discussion and Future Work}

\subsection{Novelty and Theoretical Contribution}
Most existing frameworks for temporal language modeling focus on either 
(1) retraining embeddings for each time slice \cite{hamilton2016diachronic,bamler2017dynamic}, 
or (2) employing recurrent or attention-based dynamic graph neural networks to propagate node states through time \cite{nicholson2023semantic,kazemi2020dynamic}. 
While these approaches successfully capture temporal variation, they typically incur higher computational cost, reduced interpretability, and limited ability to isolate the specific contribution of drift to the evolving network structure.

In contrast, DATGR introduces an edge-centric paradigm for temporal adaptation, conceptually related to prior edge-rewiring approaches such as DiffWire~\cite{arnaiz2022diffwire}, but extended to incorporate temporal drift signals.
Rather than recalculating entire node embeddings, DATGR explicitly updates the edge weights in the co-occurrence graph as a closed-form function of estimated drift.  
This shift from node-level to edge-level adaptation forms the central novelty of the framework, enabling the model to reflect emerging or diminishing associations between biomedical terms without relearning the entire semantic space.  
As a result, semantic evolution is encoded directly in graph structure rather than distributed across latent embedding dimensions, making individual changes traceable and auditable.

Another notable aspect lies in the logistic update rule, which balances historical persistence with drift-driven innovation.  
Whereas most temporal embedding models rely on additive or linear updates that treat all changes as equally significant, the logistic function in DATGR introduces a saturating non-linearity that dampens noise while emphasizing strong, consistent shifts in co-occurrence patterns.  
This formulation provides an intuitive regularization effect: large, stable drifts dominate small, transient fluctuations.

\subsection{Empirical and Practical Significance}
The experimental results indicate that DATGR effectively captures semantic drift while maintaining precision.  
Across all temporal windows, DATGR improves AUROC by approximately 0.066 compared with the static baseline, demonstrating stronger discriminative capability in identifying emerging relations.  
AUPRC values remain comparable to the baseline and slightly surpass it in later windows, indicating that recall increases without degrading precision.

This balance is particularly valuable for biomedical systems where false associations can propagate misinformation.  
The ability to achieve AUROC gains while preserving AUPRC parity suggests that DATGR's rewiring introduces meaningful structural corrections rather than indiscriminate connectivity expansion.  
The edge-edit patterns further reveal steady, controlled adaptation, where each temporal transition modifies only a modest subset of edges, sufficient to capture semantic novelty while preserving network coherence.

From an operational standpoint, this represents an attractive trade-off for large-scale biomedical knowledge platforms such as PubMed, bioRxiv, or ontology-curation systems.  
DATGR can be inserted as a modular update layer between corpus ingestion and embedding generation, requiring no retraining of downstream encoders. 
Because the algorithm scales linearly with the number of edges $O(|E|)$ and avoids backpropagation, it remains computationally feasible for large vocabularies.
Recent benchmarks such as Know2BIO~\cite{know2bio2023} further highlight the importance of evaluating evolving biomedical knowledge graphs under temporal dynamics -- reinforcing the practical value of drift-aware approaches like DATGR for real-world corpora.

Its modular design also makes it well-suited for retrieval-augmented generation (RAG) pipelines~\cite{jeong2024adaptiverag}, where maintaining semantically coherent retrieval vectors over time is crucial for factual accuracy.  
DATGR can function as a drift-regularization layer, helping ensure that evolving biomedical concepts remain temporally consistent without retraining large encoder models.

\subsection{Comparison with Previous Paradigms}
Table~\ref{tab:comparison} conceptually contrasts DATGR with representative temporal modeling paradigms.

\begin{table}[!t]
    \caption{Comparison Between DATGR and Prior Temporal Models.}
    \label{tab:comparison}
    \centering
    \setlength{\tabcolsep}{3pt}
    \renewcommand{\arraystretch}{1.05}
    \begin{tabular}{|c|c|c|c|}
    \hline
    \textbf{Model} & \textbf{Target} & \textbf{Retrain.} & \textbf{Interp.} \\
    \hline
    Static Embed. (w2v, GloVe) & None & No & High \\
    Dyn. Embed. (Temp2Vec) & Node vec. & Yes & Mod. \\
    Dyn. GNN & Node states & Yes & Low \\
    \textbf{DATGR (ours)} & Edge weights & No & High \\
    \hline
    \end{tabular}
    \vspace{2pt}
    
    \footnotesize{\textit{Abbreviations:} Dyn.=Dynamic; Embed.=Embedding; Retrain.=Retraining; Interp.=Interpretability; Mod.=Moderate.}
    \end{table}

As shown, DATGR provides an interpretable, retraining-free mechanism that captures temporal sensitivity at the relational level.  
It complements rather than replaces neural or embedding-based paradigms, serving as a stable substrate upon which expressive temporal GNNs or continual learning models can build.

\subsection{Interpretability and Theoretical Perspective}
Beyond quantitative metrics, DATGR's rewiring dynamics yield a transparent lens into biomedical language evolution.  
Each edge weight explicitly reflects the co-occurrence strength between two terms and evolves according to measurable semantic drift.  
Visualizing these rewiring trajectories reveals how associations such as ``mRNA vaccine'' intensified after 2020 as the term displaced the earlier, broader ``RNA vaccine'' usage, or how ``viral vector'' co-occurrences waned as mRNA-based delivery methods rose to prominence.  
This explicit traceability, linking edge changes directly to drift magnitudes, offers interpretability rarely achievable with latent temporal embeddings or neural recurrent models.
\subsection{Future Research Directions}
Future work can extend DATGR along several methodological and application axes:

\begin{enumerate}
    \item \textbf{Learnable Parameterization:}  
    Introduce adaptive or learnable coefficients $(b_0,b_1,b_2,b_3)$ so the update rule can calibrate itself dynamically to different domains or drift intensities, effectively creating a ``drift-aware optimizer.''

    \item \textbf{Multimodal Drift Integration:}  
    Combine textual drift with external modalities such as citation networks or MeSH ontologies to distinguish linguistic change from topic diffusion.

    \item \textbf{Hybridization with Neural Temporal Models:}  
    Use DATGR as a structural prior for temporal GNNs, where rewired edges guide attention toward semantically active relationships while retaining interpretability.

    \item \textbf{Evaluation on True Chronological Corpora:}  
    Apply DATGR to genuine temporal datasets (e.g., PubMed 2000--2024, COVID-19 literature) to evaluate its ability to track known paradigm shifts in biomedical science, following recent advances in temporal biomedical knowledge graph modeling~\cite{postiglione2024tkg}.

    \item \textbf{Continual RAG Integration:}  
    Integrate DATGR within retrieval-augmented generation pipelines to maintain synchronization between evolving knowledge graphs and static LLM representations.
\end{enumerate}

\subsection{Broader Implications}
DATGR demonstrates that meaningful temporal adaptation can be achieved without deep retraining or opaque neural recurrence.  
By encoding drift as explicit graph structure rather than opaque neural state, it makes temporal adaptation both auditable and domain-transferable.
In an era of rapidly evolving biomedical knowledge and large-scale AI discovery, such transparent mechanisms offer a path toward sustainable, continually updating scientific knowledge graphs that balance accuracy, explainability, and efficiency.
\section{Conclusion}
This paper introduced DATGR, a Drift-Aware Temporal Graph Rewiring framework that captures biomedical semantic evolution through lightweight, interpretable edge updates rather than full embedding retraining. By integrating embedding-based drift estimation with a logistic rewiring rule, DATGR achieves consistent AUROC gains of approximately 0.066 over a static baseline while maintaining comparable precision-recall performance, demonstrating that edge-level adaptation effectively encodes temporal semantic change at substantially lower computational cost than dynamic GNNs or sequential fine-tuning approaches.

More broadly, DATGR illustrates that meaningful temporal adaptation need not rely on opaque neural recurrence or expensive retraining. Its transparent, graph-based update mechanism makes it well-suited for deployment in knowledge graph curation, biomedical RAG pipelines, and semantic retrieval systems where both interpretability and timeliness are critical. In doing so, it offers a principled bridge between symbolic and neural paradigms for evolving scientific text.

\section*{Acknowledgment}
The authors thank Jacqueline M. Ring, Vice Chancellor and Chief Institutional Research Officer at Rowan University, for her support and guidance throughout this work.
Portions of this manuscript were refined using OpenAI's ChatGPT to improve clarity and formatting~\cite{openai2025chatgpt}. All conceptual content, data analysis, and interpretation were developed by the authors, in compliance with IEEE's guidance on AI-generated text disclosure.


\section*{Code Availability}
The implementation and evaluation scripts for DATGR are publicly available at: \\
\url{https://github.com/sahanakv01123/DATGR}. 
The repository includes the full Colab notebook, experiment pipeline, and figure-generation scripts necessary to reproduce the reported results.

\section*{Published Version}
This paper has been accepted and published at the 2026 IEEE Conference on 
Artificial Intelligence (CAI 2026) ~\cite{vijayakumar2026datgr}. The final published version is available 
at: \url{https://doi.org/10.1109/CAI68641.2026.11536228}


\end{document}